\documentstyle[11pt,a4,epsf]{article}

\title{Neural Network Methods
for Boundary Value Problems Defined in Arbitrarily
Shaped Domains}
\author{I. E. Lagaris\thanks{Permanent address: 
Dept. of Computer Science,
University of Ioannina,  
45110 Ioannina - GREECE} \\
School of Nuclear Engineering, Purdue University \\
West Lafayette, INDIANA 47907 - USA \\ \\
A. Likas \\
Department of Computer Science, University of Ioannina \\
45110 Ioannina - GREECE \\ \\
D. G. Papageorgiou \\
Network Operations Center, University of Ioannina \\
45110 Ioannina - GREECE}

\date{}

\begin{document}

\baselineskip 18pt

\maketitle

\begin{abstract}
Partial differential equations (PDEs) with Dirichlet boundary conditions
defined on boundaries with simple geomerty
have been succesfuly
treated using sigmoidal multilayer 
perceptrons in previous works \cite{LLF1,LLF2}.
This article deals with the case of complex boundary geometry,
where the boundary is determined by a number of points that belong to it
and are closely located, so as to offer a reasonable representation.
Two networks are employed: a multilayer perceptron and a radial
basis function network. 
The later is used to account for the satisfaction of the boundary conditions.
The method has been succesfuly tested on two-dimensional and
three-dimensional PDEs and has yielded accurate solutions.   
\end{abstract}

\section{Introduction}
Neural Networks have been employed before to solve boundary
and initial value problems \cite{LLF1} as well as eigenvalue
problems \cite{LLF2}. The cases treated in the above mentioned
articles were for simple finite or extended to infinity  
{\em orthogonal box} boundaries.
However when one deals with realistic problems, as for instance
in modelling the human head-neck system \cite{FCM}  or the flow
and mass transfer in chemical vapor deposition reactors \cite{FOT},
the boundary is highly irregular and cannot be described in
terms of simple geometrical shapes, that in turn would have
allowed for a simple modelling scheme.

In this article we propose a method capable of dealing with
such kind of arbitrarily shaped boundaries.
As before \cite{LLF1,LLF2},
our approach is based on the use of feedforward artificial neural
networks (ANNs) whose approximation capabilities have been
widely aknowledged \cite{HOR,LES}. More specifically, the proposed
approach is based on the synergy of two feedforward ANNs of different
types: a multilayer perceptron (MLP) as the basic approximation element
and a radial basis function (RBF) network for satisfying the 
BCs, at the selected boundary points. In addition, our approach 
relies 
on the availability of efficient software for multidimensional 
minimization \cite{MERLIN} that is used for adjusting the parameters 
of the networks.

A solution to differential equation problems based on ANNs
exhibits several desirable features:
\begin{itemize}
\item Differentiable closed analytic form.
\item Superior interpolation properties.
\item Small number of parameters.
\item Implementable on existing specialized hardware (neuroprocessors).
\item Also efficiently implementable on parallel computers.
\end{itemize}

In the next section we describe the proposed method and derive
useful formulas, while 
in section 3, we discuss implementation procedures and
numerical techniques.
In section 4 we illustrate the method by means of examples
and we compare our results to analytically known ones. 
Finally section 5 contains conclusions and directions for future
research.

%
\section{Description of the method}
We will examine  PDEs  of the form
\begin{equation}
L\Psi = f 
\end{equation}
where $L$ is a differential operator and $\Psi = \Psi(x)$ 
($ x \in D \subset R^{(n)}$), with Dirichlet B.C.s, i.e. with
$\Psi$ being specified on the boundary $B = \partial D$.
The boundary can be any arbitrarily complex geometrical shape.
We consider that the boundary 
is defined as a set of points that are chosen so as to represent
its shape reasonably accurate. Suppose that $M$ points 
$R_1,R_2,\ldots,R_M \in B$ are chosen
to represent the boundary and hence the boundary conditions
are given by:
\begin{equation}
\Psi(R_i) = b_i, \ \ i=1,2,\ldots,M
\end{equation}

To obtain a solution to the above differential equation, the {\em
collocation} method \cite{KIN91} 
is adopted which assumes the discretization of the
domain $D$ into a set of points $\hat{D}$ (these points are denoted by
$r_i$, $i=1, \ldots, K$). The problem is then transformed into the
following system of equations:
\begin{equation}
L\Psi(r_i)=f(r_i), \forall r_i \in \hat{D}, \hbox{\ \ and\ \ } 
\Psi(R_i) = b_i, \ \forall R_i \in B
\end{equation}

Let $\Psi_M(x,p)$ denote a trial solution to the above problem
where $p$ stands for a set of model parameters to be adjusted. In
this way, the problem is transformed into the following
constrained minimization problem:
\begin{equation}
min_{p} E(p)=\sum_{i=1}^K (L\Psi_M(r_i,p) - f(r_i))^2
\end{equation}
subject to the constraints imposed by the B.Cs 
\begin{equation}
\Psi_M(R_i,p) = b_i, \ \ i=1,2,\ldots,M
\end{equation}
The above constrained optimization problem may be tackled 
in a number of ways.
\begin{enumerate}
\item Either devise a model $\Psi_M(r,p)$, 
such that the constraints are exactly satisfied 
by construction and hence use unconstrained optimization techniques,
\item  Or, use a suitable constrained optimization method
for non-linear constraints.
For instance: Lagrange Multipliers, Active Set methods or a Penalty 
Function approach.
\end{enumerate}
A model suitable for the first approach is a synergy 
of two feedforward neural networks of different type, 
and it can be written as:
\begin{equation}
\Psi_M(x,p) = N(x,p) + \sum_{l=1}^M q_l e^{-\lambda |x-R_l|^2}
\end{equation}
where $N(x,p)$ is a multilayer perceptron (MLP) 
with the weights and biases
collectively denoted by the vector $p$. The sum in the above equation 
is an RBF network with $M$ hidden units that all share
a common exponential factor $\lambda$. 

For a given set of MLP parameters $p$, the coefficients
$q_l$ are uniquely determined by requiring that the boundary 
conditions are satisfied, ie:
\begin{equation}
b_i - N(R_i,p) = \sum_{l=1}^M q_l e^{-\lambda |R_i-R_l|^2}
\ \ \ i=1,2,\ldots,M
\end{equation}
Namely one has to solve a linear system,
$ Aq = c$, where $A_{ij}=e^{-\lambda |R_i-R_j|^2}$ and $c_i=b_i-N(R_i,p)$
where $i,j = 1,\ldots,M$.

We consider now a penalty function method to solve the constrained
optimization problem. 
The model in this case is simply $ \Psi_M(x,p)=N(x,p)$.
The error function to be minimized is now given by:
\begin{equation}
E(p,\eta)=\sum_{i=1}^K (L N(r_i,p) - f(r_i))^2 + \eta\sum_{i=1}^M
(\Psi_M(R_i)-b_i)^2
\end{equation}
where the penalty factor $\eta$, takes on higher and higher positive 
values depending on how accurately the BCs are to be satisfied. 

The MLP-RBF synergy satisfies exactly the BCs but it is slow.
At every evaluation of the model one needs to solve a linear system
which may be quite large, depending on the problem. Also since many
 efficient optimization methods need the gradient of the error function,
one has to solve for each gradient component an aditional linear system
of the same order. This makes the process computationally intensive.
On the other hand, the penalty method is very efficient, however satisfies
the BCs approximately only.
In practice a combination of these two methods may be used profitably in 
the following manner. 
\begin{itemize}
\item 
Use the penalty method to obtain a reasonable model that satisfies
to some extend the BCs. 
\item
Improve the model, using for a few iterations the synergy method,  
that will in addition satisfy the BCs exactly.
\end{itemize}
We used the above combination in all of our experiments and our results
are quite encouraging.

%
\section{Implementation and Numerical techniques}

The MLPs we have considered contain one hidden layer 
with sigmoidal hidden units and a linear output
that is computed as:
\begin{equation}
N(x,p) = \sum_{i=1}^H v_i \sigma(\sum_{j=1}^n w_{ij}x_j +u_i)
\end{equation}
where $n$ is the number of input units,
$H$ is the number of the hidden units and $\sigma(z)=[1+e^{-z}]^{-1}$.

In order to minimize the error $E(p)$, optimization techniques are employed
that require the computation of the derivatives 
$\frac{\partial E}{\partial p}$ and, consequently, the derivatives
$\frac{\partial\Psi_M}{\partial p}$
which are listed below:
\begin{equation}
\frac{\partial \Psi_M(x,p)}{\partial p}  =
\frac{\partial N(x,p)}{\partial p} +
\sum_{l=1}^M \frac{\partial q_l}{\partial p} e^{-\lambda (x-R_l)^2}
\end{equation}
Since $q_l = \sum_{i=1,M} A_{li}^{-1}(b_i-N(R_i,p)) $
we get:
$$
\frac{\partial q_l}{\partial p} =
  -\sum_{i=1}^M A_{li}^{-1}\frac{\partial N(R_i,p)}{\partial p}
$$
i.e. one has to solve as many $M \times M$ linear systems as the number
of the parameters $p$.
Derivatives of the MLP with respect to either the parameters
$p$ or the input variables can be easily derived and are given
in \cite{LLF1}.

In order to apply the proposed method, first the value of
$\lambda$ must be specified that defines the linear system
(matrix $A$).
In our experiments the linear system was solved using standard
Choleski decomposition for the matrix A. We did not use special
methods for sparse linear systems nor any parallel programming techniques.

For large values of $\lambda$
the Gaussian terms in the RBF are all highly localized so that affect 
the model only in the neighborhood of the boundary points. 
In other words the RBF contributes a ``correction'' to account for the BCs.
For small values of $\lambda$, the matrix looses rank and becomes
singular. So  $\lambda$ must be selected with caution.
A good choice is found to be: $\lambda \approx \frac{1}{a^2}$,
where $a$ is the minimum distance between any two points on the boundary, ie:
$ a =min_{i,j}[|R_i-R_j|] $, where $i,j =1,.2,\cdots,M$.
Note that different $\lambda$'s may also be used instead of a common one in 
equation (6). In that case the corresponding $a_j$ would be the distance
of the closest boundary neighbouri to point $R_j$, i.e. 
$a_j = min_i[|R_i-R_j|] $, where $ i = 1,.2,\cdots,M$.
However a common $\lambda$ leads to a symmetric matrix $A$ that in turn 
renders the linear system easier to solve.

Training of the MLP network so as to minimize the error of eq. (4)
can be accomblished using any minimization procedure such as
gradient descent (backpropagation or any of its variants),
conjugate gradient, Newton methods etc. Many effective
minimization techniques are provided by the Merlin/MCL multidimensional
optimization system \cite{MERLIN,MCL} which has been employed in
our experiments.
It has been earlier demonstrated \cite{KARRAS,LIKAS}, that
an improvement in the generalization of the neural model is achieved,
if the sigmoidal parameters are kept inside a limited range 
such that the exponentials do not loose precision.
Hence box-constrained optimization techniques should be used to
guarantee the above requirement.
From the variety of the minimization methods offered by
the Merlin optimization environment, the (quadratically convergent)
BFGS method \cite{BFGS} seemed to have the best performance.

When solving  problems requiring several hundreds of boundary points 
(and thousands of domain points) the method may become relatively slow. 
There are several techniques that may be applied in order to accelerate 
the process. The linear systems are sparse and hence one can employ
iterative sparse solvers instead of the Choleski factorization method
that we used here.
When computing the gradient of the error function, one has to solve
many linear systems with identical left hand sides and hence one may 
use special methods that currently are under investigation and development \cite{Gallo}.
Parallel programming techniques for machines with many cpus are also applicable.
The most efficient implementation however would be one that will utilize 
specialized hardware (neuroprocessors).

We describe now the strategy followed in detail.
\begin{enumerate}
\item
At first we use the efficient penalty function approach 
(with $\eta=100$ in all tests)  to obtain an MLP
network that approximates the solution both inside the domain
and on the boundary.
\item
Then we switch to the MLP-RBF method with initial parameter
values for the MLP network those obtained from the penalty method. 
Therefore the MLP-RBF method starts from a low error value and requires
only a few minimization steps in order to reach a solution of even lower
error value which in addition satisfies the BCs exactly.
\end{enumerate}

\section{Examples}

\subsection{Two Dimensional Problems}

{\bf Problem 1:}
Consider the problem:
\begin{equation}
\nabla^2 \Psi(x,y) = e^{-x}(x-2+y^3+6y), \ \ x,y \in [0,1]
\end{equation}
with boundary conditions:
$$
\Psi(0,y) = y^3, \Psi(1,y) = \frac{1+y^3}{e},
\Psi(x,0) = x e^{-x}, \Psi(x,1) = e^{-x}(1+x)
$$
\begin{figure}
\centerline{\epsfysize=10cm\epsfxsize=14cm\epsffile{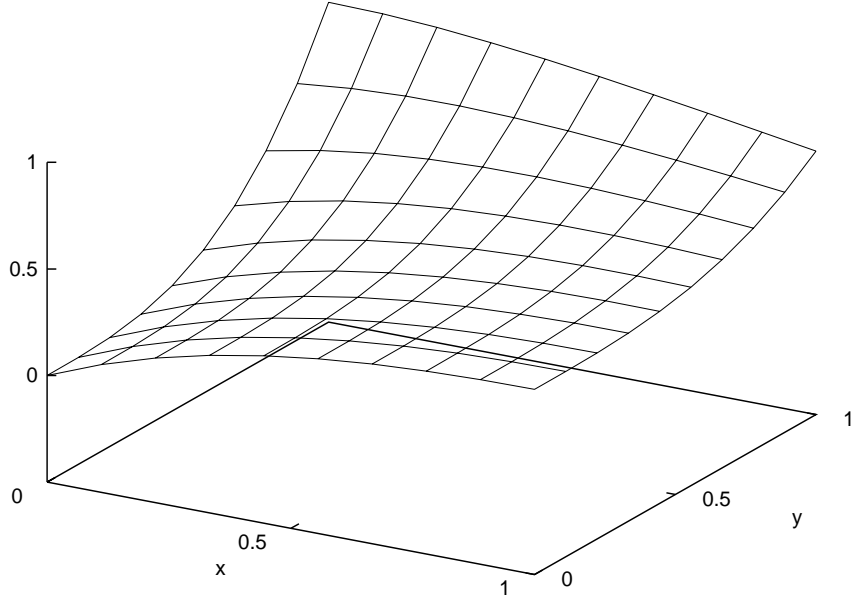}}
\caption{Exact solution of Problem 1}
\end{figure}

\begin{figure}
\centerline{\epsfysize=10cm\epsfxsize=14cm\epsffile{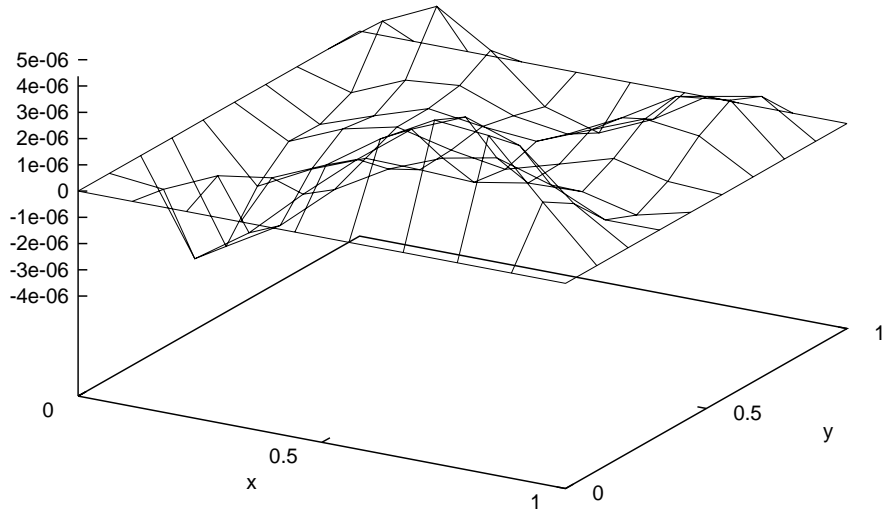}}
\caption{Accuracy of obtained solution to problem 1}
\end{figure}

The analytic solution is: $\Psi_a(x,y)=e^{-x}(x+y^3) $.
This example has also been treated in \cite{LLF1}.
Here the problem is treated by picking points on the boundary
as if it were any arbitrary shape.
More specifically, we take the following points $(x,y)$ on the boundary,
where $m_x$ and $m_y$ denote the number of points which 
divide the interval $[0,1]$ on the x-axis and y-axis respectively
and $\delta_x=1/(m_x-1)$, $\delta_y=1/(m_y-1)$:
$$
((i-1)\delta_x,0), \ \ i=1, \ldots, m_x-1
$$
$$
((i-1)\delta_x,1), \ \ i=1, \ldots, m_x-1
$$
$$
(0,(i-1)\delta_y), \ \ i=1, \ldots, m_y-1
$$
$$
(1,(i-1)\delta_y), \ \ i=1, \ldots, m_y-1
$$
After this selection, a test is made to remove duplicates, which in
this case are the points at the corners of the rectangle boundary. 
In our experiments we have considered $m_x=m_y=m=10$ and, therefore, 
the total number of points taken on the boundary is $ M = 4(m-1)=36 $.
For the points inside the definition domain we pick points
on a rectangular grid by subdividing the $ [0,1] $ interval in 
10 equal subintervals that correspond to 9 points in each direction.
Thus a total of $K=81$ points are selected. The analytic solution is
presented in Fig. 1, while the accuracy $|\Psi_M(x,y)-\Psi_a(x,y)|$ of the
obtained solution is presented in Fig. 2. In all two-dimensional examples
we used an MLP with 20 hidden units.

{\bf Problem 2:}

\begin{figure}
\centerline{\epsfysize=10cm\epsfxsize=14cm\epsffile{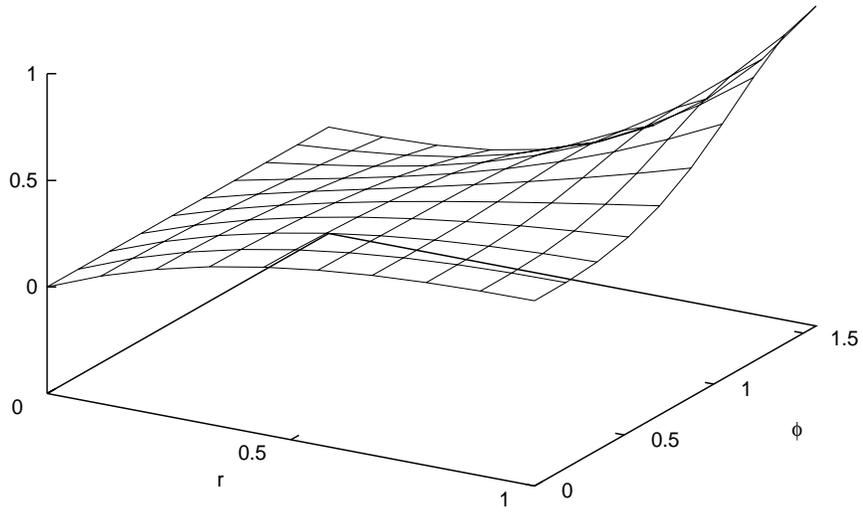}}
\caption{Exact solution of problem 2}
\end{figure}

\begin{figure}
\centerline{\epsfysize=10cm\epsfxsize=14cm\epsffile{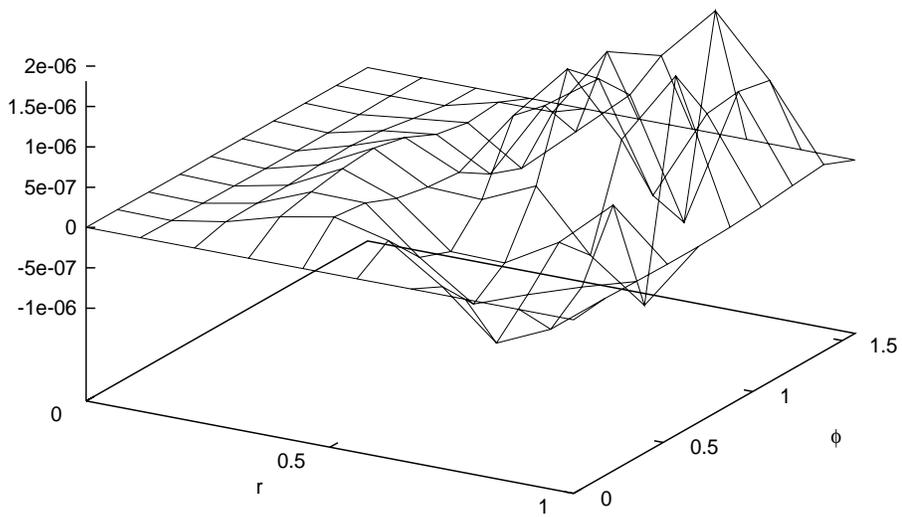}}
\caption{Accuracy of obtained solution to problem 2}
\end{figure}

The same problem is solved with 
the boundary being the first quarter of the unit circle. 
The solution domain is defined as the rectangle $[0,1]\times [0,\pi/2]$
on the polar coordinates $(r,\phi)$.
To obtain the boundary points $(x,y)$ we first defined the
boundary points $(r,\phi)$ in the polar 
coordinates (according to the procedure
of the previous problem) and then we computed  the $(x,y)$ values:
$x=rcos\phi$, $y=rsin\phi$. We have used $M=37$ boundary points
and  $K=81$ grid points.  The exact solution and the
accuracy of the obtained solution are 
displayed in Fig. 3 and 4 in the $(r,\phi)$ coordinates.

{\bf Problem 3:}

\begin{figure}
\centerline{\epsfysize=10cm\epsfxsize=14cm\epsffile{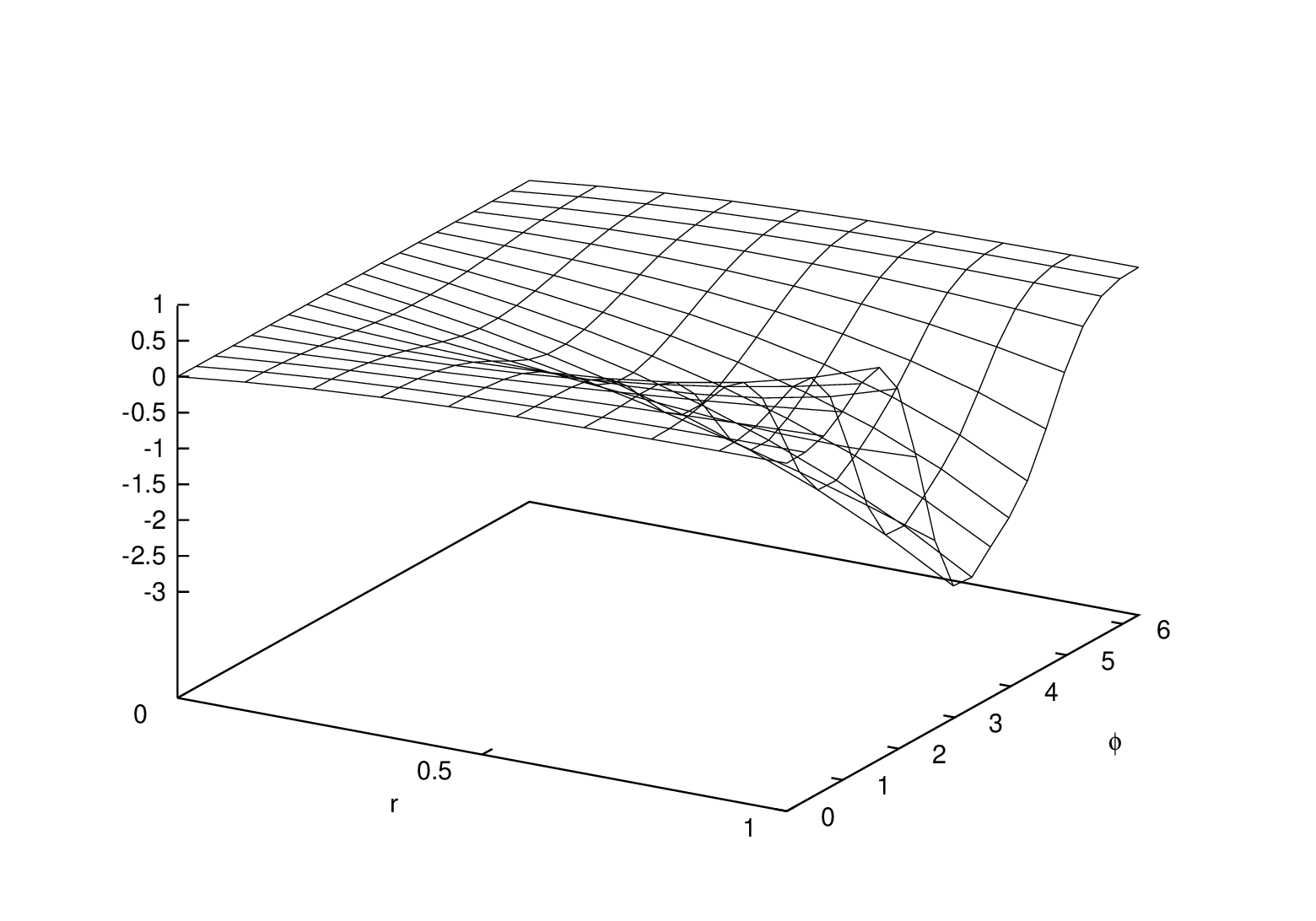}}
\caption{Exact solution of problem 3.}
\end{figure}

\begin{figure}
\centerline{\epsfysize=10cm\epsfxsize=14cm\epsffile{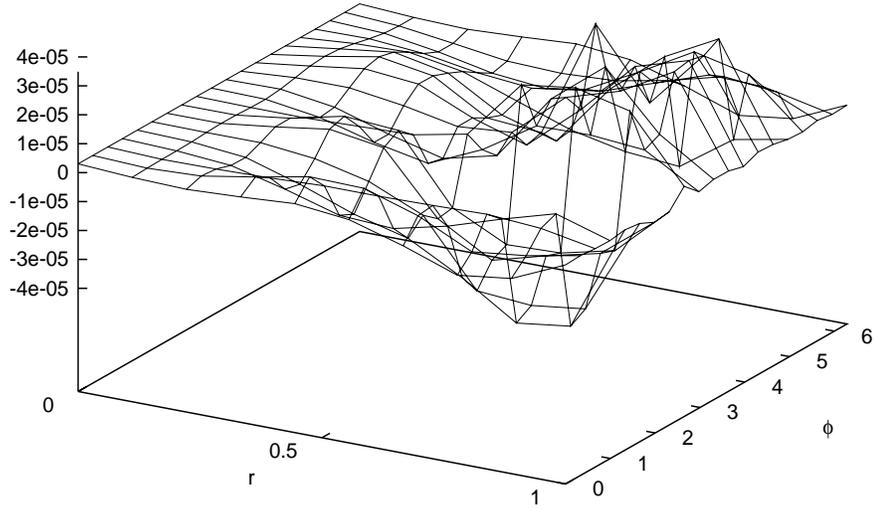}}
\caption{Accuracy of obtained solution to problem 3.}
\end{figure}

Finally we solved eq. (11) 
when the boundary is the unit circle.
The solution domain is defined as the rectangle $[0,1]\times [0,2\pi]$
on the polar coordinates $(r,\phi)$.
The boundary points $(x,y)$ are defined as 
$x=rcos\phi_i$, $y=rsin\phi_i$, where $\phi_i=2\pi i/m$,
$i=1,\ldots, m-1$.
We have used $M=20$ boundary points
and  $K=153$ grid points.
The analytic solution and the
accuracy of the obtained solution are 
displayed in Fig. 5 and 6 respectively in the $(r,\phi)$ coordinates.    

\subsection{Three Dimensional Problems}

{\bf Problem 4:}
Consider the problem:
\begin{equation}
\nabla^2 \Psi(x,y,z) = e^{x}y^2+ z^2 siny, \ \ x,y,z \in [0,1]
\end{equation}
with 
analytic solution: $\Psi_a(x,y)=e^{x} y^2 +(z^2-2)siny$ known at the
boundary.

\begin{figure}
\centerline{\epsfysize=10cm\epsfxsize=14cm\epsffile{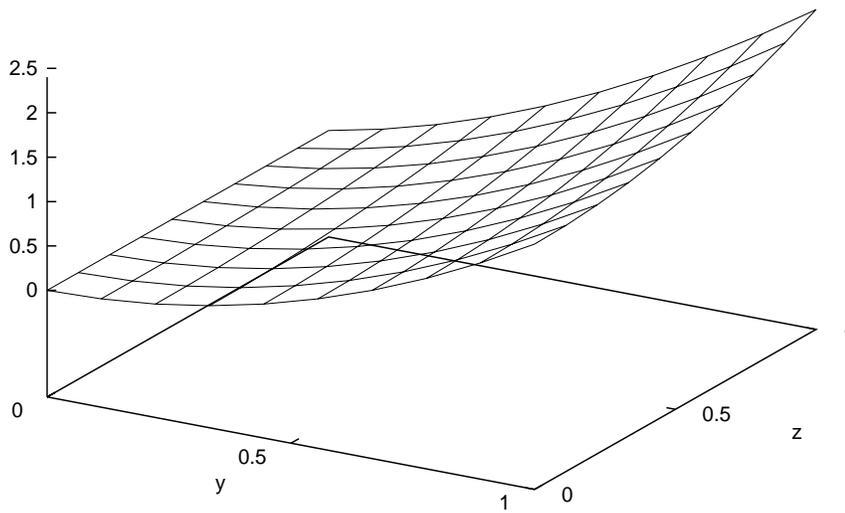}}
\caption{Exact solution of problem 4 for $x=0.5$.} 
\end{figure}

\begin{figure}
\centerline{\epsfysize=10cm\epsfxsize=14cm\epsffile{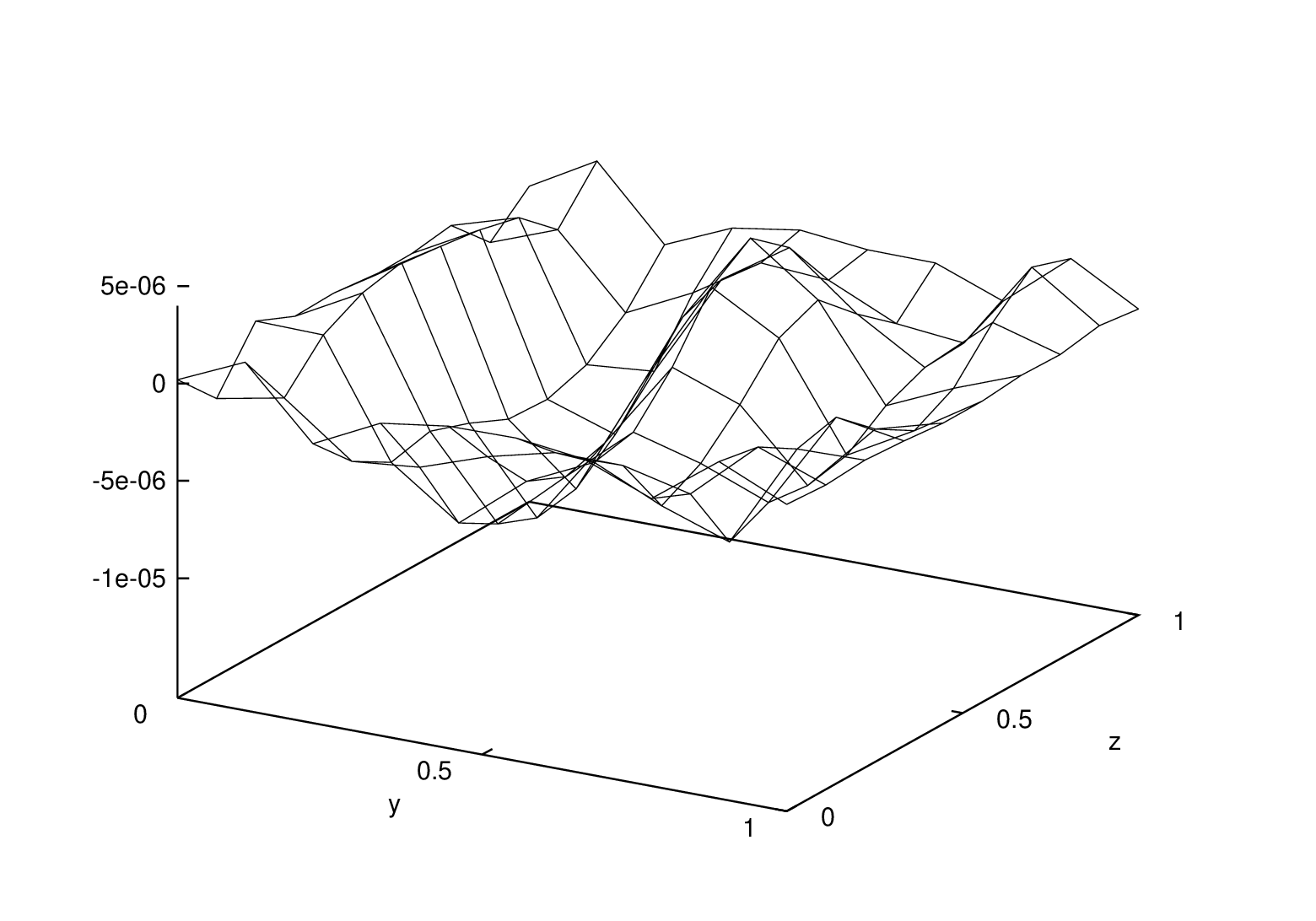}}
\caption{Accuracy of obtained solution to problem 4 for $x=0.5$.}
\end{figure}

Similarly with the approach described in Problem 1, we define
the boundary points
by dividing the $[0,1]$
interval along the x-axis, y-axis, z-axis with $m_x=m_y=m_z=m=7$ points
respectively  and taking the points $(x,y,z)$:  
$$
((i-1)\delta,(i-1)\delta,0), \ \ i=1, \ldots, m-1
$$
$$
((i-1)\delta,(i-1)\delta,1), \ \ i=1, \ldots, m-1
$$
$$
((i-1)\delta,0,(i-1)\delta), \ \ i=1, \ldots, m-1
$$
$$
((i-1)\delta,1,(i-1)\delta), \ \ i=1, \ldots, m-1
$$
$$
(0,(i-1)\delta,(i-1)\delta), \ \ i=1, \ldots, m-1
$$
$$
(1,(i-1)\delta,(i-1)\delta), \ \ i=1, \ldots, m-1
$$
where $\delta=1/(m-1)$.

After this specification a test is made to remove duplicates
or points that were very close to another point, 
and the final number of boundary points was $M=218$. 
For the points inside the definition domain we pick points
on a rectangular grid subdividing the $ [0,1] $ interval in
10 equal subintervals that correspond to 9 points in each direction
defining a total of $K=729$ points. The analytic solution for ($x=0.5)$
is presented in Fig. 7, while the accuracy $|\Psi_M(0.5,y,z)
-\Psi_a(0.5,y,z)|$ of the
obtained solution is presented in Fig. 8. In all three-dimensional examples
we used an MLP with 40 hidden units.

{\bf Problem 5:}
We considered the previous problem:
\begin{equation}
\nabla^2 \Psi(x,y,z) = e^{x}y^2+ z^2 siny 
\end{equation}
on the domain $[0.5,1]\times [0,\pi/2] \times [0,\pi/2]$ on the
spherical coordinates $(r,\phi,\theta)$.
Similarly with the approach described in Problem 4, we define
the boundary points
by dividing the intervals 
r-axis, $\phi$-axis, $\theta$-axis with $m=7$ points
respectively  and taking the points $(r,\phi,\theta)$:
$$
((i-1)\delta_r,(i-1)\delta_\phi,0), \ \ i=1, \ldots, m-1
$$
$$
((i-1)\delta_r,(i-1)\delta_\phi,\pi /2), \ \ i=1, \ldots, m-1
$$
$$
((i-1)\delta_r,0,(i-1)\delta_{\theta}), \ \ i=1, \ldots, m-1
$$
$$
((i-1)\delta_r,\pi/2,(i-1)\delta_{\theta}), \ \ i=1, \ldots, m-1
$$
$$
(0.5,(i-1)\delta_\phi,(i-1)\delta_\theta), \ \ i=1, \ldots, m-1
$$
$$
(1,(i-1)\delta,(i-1)\delta_\theta), \ \ i=1, \ldots, m-1
$$
where $\delta_r=0.5/(m-1)$, $\delta_\phi=0.5/(m-1)$, $\delta_\theta=0.5/(m-1)$.
From the $(r,\phi, \theta)$ values we obtained the corresponding
$(x,y,z)$ points using the well-known transformation.

\begin{figure}
\centerline{\epsfysize=10cm\epsfxsize=14cm\epsffile{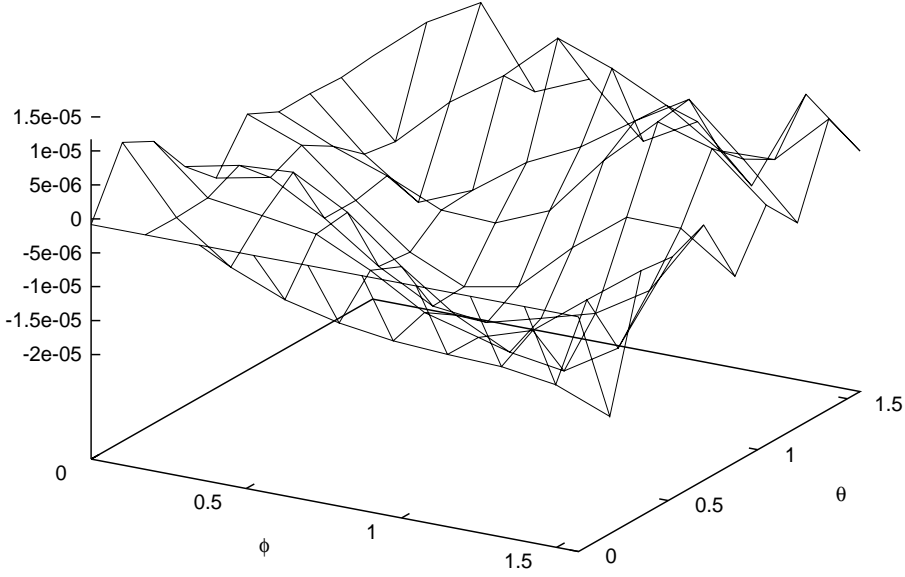}}
\caption{Exact solution of problem 5 for $r=0.75$.}
\end{figure}

\begin{figure}
\centerline{\epsfysize=10cm\epsfxsize=14cm\epsffile{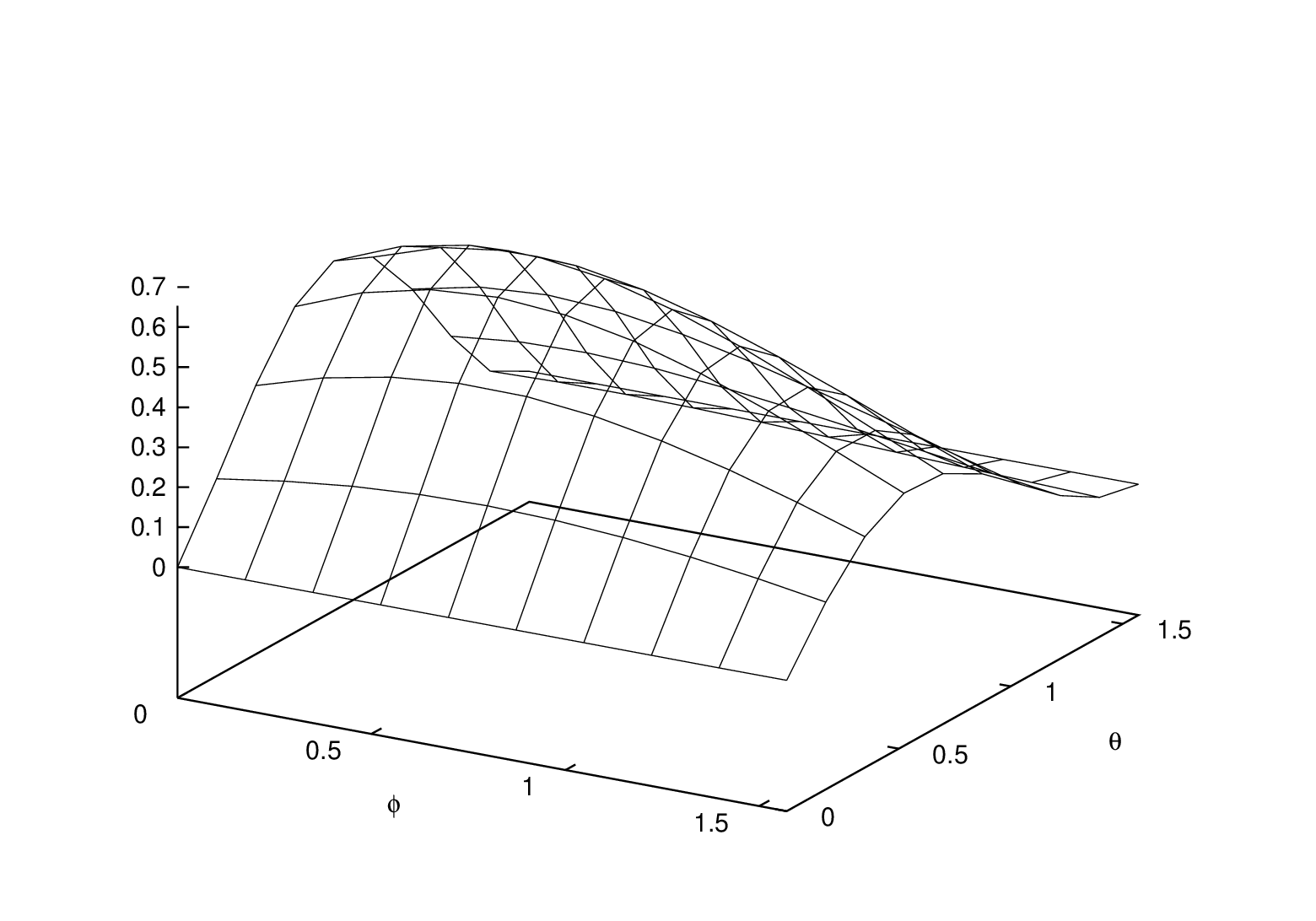}}
\caption{Accuracy of obtained solution to problem 5 for $r=0.75$.}
\end{figure}

After this specification a test is made to remove duplicates
or points that were very close to another point,
and the final number of boundary points was $M=176$.
For the points inside the definition domain we pick points $(r,\theta,
\phi)$
on a rectangular grid subdividing the $[0,0.5]$, $[0,\pi/2]$ intervals in
10 equal subintervals that correspond to 9 points in each direction
defining a total of $K=729$ points $(r,\phi,\theta)$. Then the $(x,y,z)$
points were obtained from the $(r,\theta,\phi)$ points. The exact
solution and accuracy 
of the obtained solution are 
displayed in Fig. 9 and 10 in the $(\phi,\theta)$ coordinates for $r=0.75$.

\section{Conclusions}
We have presented a method for solving differential equations
with Dirichlet BCs, where the boundary can be of arbitrary shape
and is discretized to obtain a set of boundary points. The method
is based on the synergy of MLP and RBF artificial neural networks 
and provides accurate and differentiable solutions (in a closed
analytic form) that exactly satisfy the BCs at the selected boundary
points. Moroeover it is possible to implement the method
on specialized hardware (neuroprocessors) to significantly improve 
the required solution time.
The proposed method is quite general and can be used for a wide class
of PDEs with Dirichlet BCs, regardless of the shape of the boundary.
The only requirement is that enough boundary points are selected
so as to represent the boundary shape with sufficient accuracy.

Future work will focus on the application of the method to
real-world 3-D problems, containing surfaces of real objects
with arbitrary shapes as boundaries. 
Interesting problems of this kind arise in many scientific fields.  
We have a strong interest in
implementing the method on both, general purpose parallel
hardware and on neuroprocessors. 
The later would reveal the full potential of the proposed approach
and would lead to the development of specialized machines,
that will allow the treatment of difficult and computationally 
demanding problems.

{\bf Ackowledgement:} One of us (I. E. L) wishes to acknowledge the warm
hospitality offered by professors Ishii and Tsoukalas of Purdue University,
at the school of Nuclear Engineering, during his stay at Lafayette.

%

\end{document}